# Automatic 2D-3D Registration without Contrast Agent during Neurovascular Interventions

Robert Homan, René van Rijsselt, and Daniel Ruijters

*Abstract*—Fusing live fluoroscopy images with a 3D rotational reconstruction of the vasculature allows to navigate endovascular devices in minimally invasive neuro-vascular treatment, while reducing the usage of harmful iodine contrast medium. The alignment of the fluoroscopy images and the 3D reconstruction is initialized using the sensor information of the X-ray C-arm geometry. Patient motion is then corrected by an image-based registration algorithm, based on a gradient difference similarity measure using digital reconstructed radiographs of the 3D reconstruction. This algorithm does not require the vessels in the fluoroscopy image to be filled with iodine contrast agent, but rather relies on gradients in the image (bone structures, sinuses) as landmark features. This paper investigates the accuracy, robustness and computation time aspects of the image-based registration algorithm. Using phantom experiments 97% of the registration attempts passed the success criterion of a residual registration error of less than 1 mm translation and 3° rotation. The paper establishes a new method for validation of 2D-3D registration without requiring changes to the clinical workflow, such as attaching fiducial markers. As a consequence, this method can be retrospectively applied to pre-existing clinical data. For clinical data experiments, 87% of the registration attempts passed the criterion of a residual translational error of < 1 mm, and 84% possessed a rotational error of < 3°.

## I. Introduction

ENDOVASCULAR interventional treatment of neurovascular lesions is performed under fluoroscopic guidance using a X-ray C-arm unit. In the fluoroscopic images the vasculature can be visualized by injecting iodine contrast agent for navigation purposes while advancing the catheter or guidewire and inspection of the vascular morphology. Particularly, two-dimensional (2D) digital subtraction angiography (DSA) is used to visualize the vessel tree, without showing the surrounding bony structures [1-4]. Unfortunately, iodine contrast agent is toxic and especially for patients with kidney failure the contrast load should be kept to a minimum. The DSA technique can be further enhanced to form a 'roadmap' by overlaying a static DSA of the vessels with a subtraction image showing the catheter and guide wire [5,6]. This roadmap technique facilitates the catheter navigation, and also helps to reduce the contrast agent load. The roadmap is, however, invalidated by any movement of the C-arm, the patient table, or changing the source-to-image distance, etc. 3D-Roadmap is a technique whereby a three-dimensional rotational angiography (3DRA) reconstruction is used instead of the DSA for roadmapping purposes [7-9]. Since the 3DRA reconstruction can be shown from any angle with any zoom factor and perspective frustum, it stays valid when any of the geometrical C-arm parameters are changed. It has proven to be valuable addition to the DSA driven roadmap in further reducing contrast agent dose, while adding 3D information to the navigation of endovascular devices.

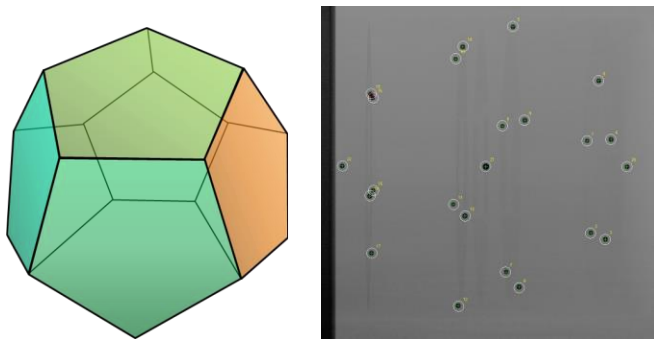

Fig. 1. The dodecahedron shape (left) is used to calibrate the C-arm pose. A led sphere is located at each vertex, which uniquely defines the position of the dodecahedron in an X-ray image (right).

The 3DRA and the live fluoroscopy image in the 3D-Roadmap technique are aligned by using the geometry sensors of the C-arm and calibration information. This delivers a very accurate alignment as long as the patient does not move. In this article we employ an algorithm that is driven by the content of the fluoroscopic image and the 3DRA dataset to additionally correct for any (rigid) patient motion.

Image-driven co-registration of 2D and 3D images has been investigated by several authors since the late 1990s [10-22]. The image-based algorithms typically take a considerable amount of time to compute, ranging from a few seconds for methods that use a model of the anatomy of interest up to a few minutes for some intensity-driven approaches [17]. Since these algorithms use the image content, sufficient landmark structures should be available in both images.

In registration methods for angiographic applications, the structures are often provided by filling the vasculature with

R. H., R. v. R., and D. R. are with the Image Guided Therapy department of Philips Healthcare, 5680DA Best, the Netherlands (e-mail: {robert.homan, rene.van.rijsselt, danny.ruijters}@philips.com).

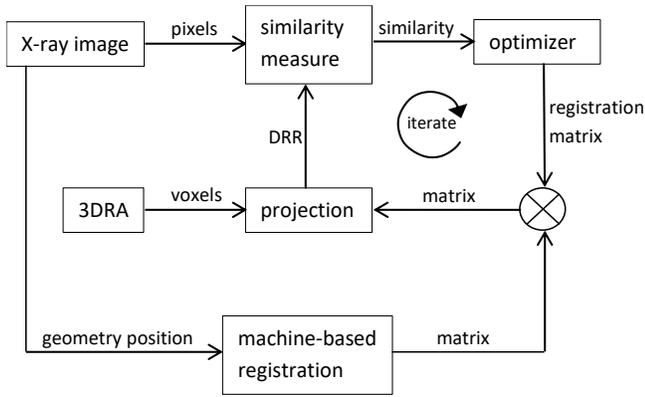

Fig. 2. Algorithm flow. The dynamic data sources during 3D roadmapping are the X-ray image and its corresponding geometry position. The 3DRA is a static data source. The projection parameters for the DRR are iteratively updated by the optimizer, while optimizing the similarity between the DRR and the X-ray image.

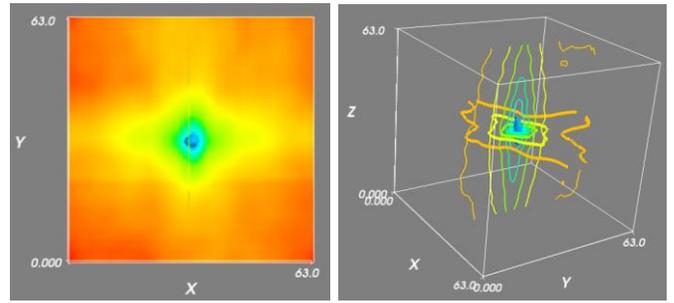

Fig. 3. Similarity as function of the displacement in the *x*- and *y*-direction (left) and additionally as function of the rotation around the *z*-axis (right). Maximum displacement is +/-15mm and +/-7.5 degrees.

iodine contrast medium [11,12,14,16,19-22]. In this paper, however, we will not use DSA images for registration, since it is one of the objectives to reduce the use of harmful contrast agent during 3D roadmapping. Most registration methods are based on a single projection, which leads to a rather large registration error for the out-of-plane translation. As long as the projection angle does not change, this is not a big hurdle as it only leads to a slight mismatch in the magnification factor between the 2D and the 3D image [19].

Mitrović *et al.* have made available clinical 2D and 3D cerebral angiographic data for validation purposes [22]. Their gold standard registration was established using external fiducial markers attached to the patient. Although it is very valuable to have data publically accessible for algorithm evaluation and comparison, the fiducial marker approach requires additional steps in the clinical workflow and cannot be performed retrospectively. Here, we introduce a new method for evaluation, using the projection images and the volumetric reconstruction of a rotational acquisition. There are several advantages to this approach; there is no adjustment to the clinical workflow required, the data can be collected retrospectively from hospital databases, and the ground truth registration is nearly perfect, since it depends only on the geometric calibration of the C-arm system and the absence of patient motion during the rotational scan.

The contributions of our paper are twofold: 1) The validation of an image-based 2D-3D registration approach that is implemented in a commercially available system and is being routinely used in clinical practice on a daily basis. 2) The introduction of a validation method using readily available clinical data, that delivers a nearly perfect ground truth. The paper is organized as follows: in Section II the applied methods are described (3D reconstruction, machine-based registration, image-based registration, and validation methods). Section III presents the results of the phantom experiments, the experiments on clinical data, and the computation time measurements. In Section IV the results and their implications and limitations are discussed, and the conclusions are summarized in Section V.

## II. METHODS AND MATERIALS

### A. 3D Reconstruction

In order to produce 3D reconstructions from rotational acquisitions, the C-arm geometry has to be calibrated to correct for varying gravitational bending of the C-arm in different orientations. The calibration is done by making a rotational acquisition of a dodecahedron phantom as presented in Fig. 1. The calibration run should possess the same properties (number of images, starting angle, etc.) as a 3D acquisition. In every X-ray image of the calibration run the position of the dodecahedron phantom can be uniquely determined, and therefore the exact position of the C-arm detector and X-ray source are known [23]. Based on the geometrical calibration the images are back-projected on the volumetric voxel grid during the reconstruction procedure [24,25].

### B. Machine-based Registration

The first step of the registration process is the machine-based co-registration, using the geometry sensors of the C-arm [26,8,9]. Since the sensor information is not corrected for gravitational bending of the C-arm, this procedure also requires calibration. The same dodecahedron phantom as in the previous section is used for this purpose. However, since now we need to calibrate any random position, a matrix of discrete rotation and angulation combinations are calibrated at intervals of 20°, and for in-between positions the calibration information is interpolated to estimate the correct pose. This delivers an accurate registration with an error of less than 0.2 mm [9], provided that the imaged subject did not move.

### C. Image-based Registration

The result of the machine-based registration step is used as initial pose for the image-based registration process. The difference between this initial position and the actual pose can only be the result of patient motion between the moment of acquiring the 3DRA data and the live fluoroscopy image. Since the machine-based registration already provides a good initialization for the image-based algorithm, there is no need for manual interaction, which is a substantial workflow

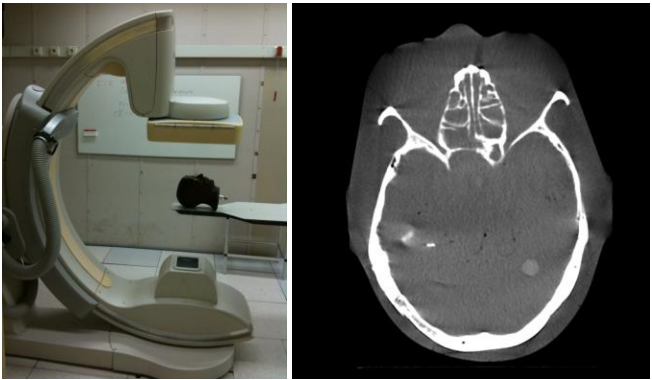

Fig. 4. The head phantom being scanned by a C-arm X-ray system (left). A slice from a cone-beam CT reconstruction of the head phantom (right).

| Id | Translation $(t_x, t_y, t_z)$ [mm] | Rotation $(r_x, r_y, r_z)$ [°] | Description |
|---|---|---|---|
| T1 | 10, 0, 0 | 0, 0, 0 | 1 cm displacement along a single axis |
| T2 | -8, 6, 0 | 0, 0, 0 | Displacement along two axes resulting in an overall displacement of 1 cm |
| T3 | 0, 0, 0 | 5, 0, 0 | A displacement by rotation around one axis resulting in an overall displacement of 1 cm |
| T4 | 4, 4, 0 | 2, 3, 4 | A composited displacement by translation and rotation resulting in an overall displacement of 1 cm |

TABLE I
INITIAL DISPLACEMENTS USED FOR PHANTOM EXPERIMENTS

The initial transformations that were imposed on the ground truth to start the image-based registration algorithm.

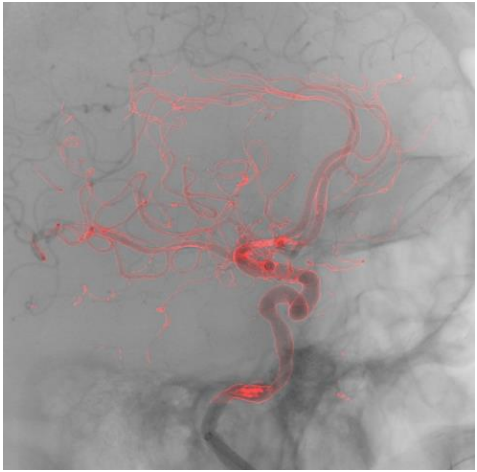

Fig. 5. Frame of a clinical rotational angiography acquisition (gray), overlaid with the outline of the vessels from its 3DRA reconstruction (red).

advantage in interventional applications [27].

The algorithmic data flow has been illustrated in Fig. 2. We use gradient difference as similarity measure, as it has proven to perform well in the 2D-3D registration of the boney anatomy [28,29]. The most computation intensive step is the generation of the digitally reconstructed radiograph (DRR) [22], and therefore we employed the GPU to accelerate this step [30]. As optimization strategy we used simulated annealing [31]. The search space of the parameters that are varied by the optimizer consists of translation in the $x$- and $y$-direction along the X-ray plane, and rotation around the $x$-, $y$-, and $z$-axis. The translation in the $z$-direction (perpendicular to the X-ray plane) is not varied, since this results only in a slight magnification variation of the 3D projection, and experiments have shown that when using only one projection plane the $z$-translation is not necessarily improved by the registration algorithm. Slices out of the 5-dimensional search space have been plotted in Fig. 3. It took about 1 hour to calculate this plot (262144 DRRs). If an exhaustive search would be conducted throughout the entire search space, it would take about 4971 hours to calculate.

The image sizes of 15 and 19 centimeters, both for the volumetric reconstructions and the overlay images, were also included in the experiments for evaluation purposes. These kind of small field of view images typically contain only very few usable landmarks for image-based registration and patient motion easily moves relevant features outside the field of view. Therefore, the commercially available software does not apply image-based registration for images of these sizes, and the results are therefore deemed not to be clinically relevant. Furthermore, overlaying a small 3D reconstruction with a larger overlay image is also excluded from the clinically relevant subset, since this scenario makes clinically little sense (the roadmap is incomplete because of the limited field of view in the 3D reconstruction), and thus is disabled in the commercial product.

### D. Validation Methods

The image-based registration results have been validated using a head phantom (Fig. 4) and clinical data (e.g., see Fig. 5). The head phantom used for the phantom experiments consisted of a human skull surrounded by poly methyl methacrylate in the shape of a human head. For these experiments 3DRA reconstructions, fluoroscopy acquisitions and exposure acquisitions were obtained, using combinations of different image formats by means of collimation. The phantom was not moved between the different acquisitions, and the initial translation and rotation was induced in the software, in order to have accurate ground truth (see Table I). The result of a registration was considered successful when the residual error after registration was less than 1 mm for the translational component and less than 3° for the rotational component.

In order to determine the accuracy and capture range of the registration using clinical data, we used 3DRA volumetric reconstructions and their raw rotational X-ray images that served to reconstruct the volumetric data. The geometrical relation between any of the raw rotational X-ray images and the 3DRA reconstruction is known very accurately and can be used as ground truth data for testing the image-based registration algorithm. The only pre-condition is that there is no patient motion during the 3DRA acquisition (about 4 sec).

The clinical data concerned intra-cranial 3DRA acquisitions during neurovascular aneurysm treatment procedures. Before feeding the images to the registration algorithm, translations up to 25 mm and rotations up to 10° were applied to the data, to create an artificial offset.

TABLE II
THE RESIDUAL OFFSET AT THE ISO-CENTER AFTER A REGISTRATION ATTEMPT IN MILLIMETERS OF THE HEAD PHANTOM, USING EXPOSURE IMAGES. R REPRESENTS THE IMAGE DIAMETER IN CENTIMETERS FOR THE 3DRA VOLUMES (VERTICAL AXIS) AND EXPOSURE IMAGES (HORIZONTAL AXIS). COLUMN T INDICATES THE INITIAL OFFSET AS GIVEN IN TABLE I. THE COLOR OF THE CELLS REFLECTS THE RESULT COMPARED TO THE CRITERION OF MAXIMALLY 1 MM RESIDUAL OFFSET (GREEN: MEETS THE CRITERION, RED: FAILED). THE BLUE OVERLAYS REPRESENT THE SUBSET THAT IS NOT CONSIDERED TO BE CLINICALLY RELEVANT.

| R | T | 15 | 19 | 22 | 27 | 31 | 37 | 42 | 48 |
|---|---|---|---|---|---|---|---|---|---|
| 15 | 1 | 0.239068701 | 0.321336241 | 0.25294446 | 0.266826926 | 10 | 10 | 10 | 10 |
|    | 2 | 0.364762781 | 0.548242396 | 0.22865607 | 0.319678461 | 0.252484378 | 10 | 10 | 10 |
|    | 3 | 0.257861007 | 0.707424205 | 0.704707387 | 0.602393802 | 0.329752744 | 0 | 0 | 0 |
|    | 4 | 0.330123083 | 0.376721064 | 0.352653598 | 0.318009371 | 5.656854249 | 5.656854249 | 5.656854249 | 5.656854249 |
| 19 | 1 | 0.284208431 | 0.207214595 | 0.301623441 | 0.269190267 | 0.276155626 | 0.237102739 | 10 | 10 |
|    | 2 | 0.113631707 | 0.287519349 | 0.30511488 | 0.167393018 | 0.25119016 | 10 | 10 | 10 |
|    | 3 | 0.130065836 | 0.323811176 | 0.19749974 | 0.301900983 | 0.864007691 | 0.202095458 | 0 | 0 |
|    | 4 | 0.315853906 | 0.370036268 | 0.997348791 | 0.78164048 | 0.205962836 | 5.656854249 | 5.656854249 | 5.656854249 |
| 22 | 1 | 10 | 1.053702146 | 0.132567307 | 0.145487531 | 0.128853624 | 10 | 10 | 10 |
|    | 2 | 0.202295724 | 10 | 0.182418639 | 0.309041162 | 10 | 10 | 10 | 10 |
|    | 3 | 0.215387984 | 0.119751254 | 0.100450692 | 0.219379691 | 0.083920503 | 0 | 0 | 0 |
|    | 4 | 5.656854249 | 5.656854249 | 5.656854249 | 5.656854249 | 5.656854249 | 5.656854249 | 5.656854249 | 5.656854249 |
| 27 | 1 | 33.06775331 | 0.128134556 | 0.092381836 | 0.286567588 | 0.149854243 | 0.157359332 | 0.199866858 | 0.32111851 |
|    | 2 | 0.186106529 | 0.200074986 | 0.227125164 | 0.237617339 | 0.172927846 | 0.243293157 | 0.159430236 | 0.433360693 |
|    | 3 | 0 | 0.038981491 | 0.057190998 | 0 | 0.322092368 | 0.12637223 | 0.030329637 | 0.112811914 |
|    | 4 | 5.656854249 | 1.41012824 | 0.044660149 | 0.16967098 | 0.074520411 | 0.106722482 | 0.138891895 | 0.101000642 |
| 31 | 1 | 0.067026369 | 0.073717577 | 0.175386167 | 0.124895466 | 0.21086983 | 0.235544607 | 0.089258014 | 0.187384404 |
|    | 2 | 0.672910211 | 0.378168489 | 0.159090075 | 0.605360843 | 0.342506084 | 0.231230098 | 0.200469605 | 0.305959428 |
|    | 3 | 0 | 0.36469413 | 0.170479709 | 0.012073218 | 0 | 0 | 0.296041199 | 0.264877358 |
|    | 4 | 0.553646537 | 0.175313 | 0.063436022 | 0.196447315 | 0.150164806 | 0.057786006 | 0.214504821 | 5.656854249 |
| 37 | 1 | 0.17130864 | 0.653824938 | 0.117196912 | 0.120575547 | 0.245827846 | 0.194299382 | 0.147573278 | 0.265413413 |
|    | 2 | 0.181108393 | 0.218625616 | 0.197451994 | 0.266659877 | 0.242089157 | 0.208976937 | 0.242276371 | 0.18443221 |
|    | 3 | 0 | 0.063020633 | 0.022003331 | 0.075297358 | 0.022369805 | 0.026376612 | 0.11372079 | 0.1094765 |
|    | 4 | 0.178833356 | 0.184523981 | 0.170308844 | 0.196999282 | 0.111265885 | 0.270169974 | 0.136859113 | 0.071869785 |
| 42 | 1 | 0.087987911 | 0.227042744 | 0.147209386 | 0.087097836 | 0.150399044 | 0.128447892 | 0.406328793 | 0.248597189 |
|    | 2 | 0.355978946 | 0.275506878 | 0.121744035 | 0.127802817 | 0.209009569 | 0.244462697 | 0.344078901 | 0.369809802 |
|    | 3 | 0.055330233 | 0 | 0.071238328 | 0.151826546 | 0.05216698 | 0.107046015 | 0.054357269 | 0.022992735 |
|    | 4 | 0.274239823 | 0.105654536 | 0.088061343 | 0.526553853 | 0.101065931 | 0.088012432 | 0.130766687 | 0.069262090 |
| 48 | 1 | 0.296620707 | 0.514653692 | 0.155071913 | 0.23454986 | 0.446570548 | 0.181106209 | 0.292766682 | 0.481124088 |
|    | 2 | 0.711045709 | 0.212679313 | 0.517874541 | 0.747803611 | 0.187598081 | 0.180614617 | 0.238636313 | 0.293108581 |
|    | 3 | 0.17429604 | 0.520258805 | 0.005166879 | 0.282007334 | 0.084395441 | 0.411678155 | 0.09700133 | 0.169570252 |
|    | 4 | 0.921812083 | 0.740408702 | 0.317822418 | 0.255842378 | 0.071648263 | 0.122301268 | 0.273067857 | 0.060249481 |

## III. RESULTS

### A. Phantom Experiments

The following list of image format diameters was used: [15, 19, 22, 27, 31, 37, 42, 48] cm for the 3DRA, fluoroscopy and exposure acquisitions. Four different initial positions were used for each data combination, as is specified in Table I. 3DRA was registered to both fluoroscopy and exposure X-ray images for each of the image format combinations. This required in total 512 registration attempts that were evaluated according the success criteria. The results regarding the offset between the registration result and the ground truth are listed in Table II for the exposure images. As can be seen in Table II, for the exposure images 84% of all registration attempts passed the defined criterion, while 97% of the registration attempts within the clinically relevant subset passed the criterion. For the fluoroscopy images the results are very similar; 83% of all attempts passed the criterion, while 97% within the clinically relevant subset passed the criterion. The distributions of the residual translation errors after registration are shown in Fig. 6 for the exposure and fluoroscopy runs regarding all runs and for the clinically relevant subset.

In order to further examine the influence of boney landmark structures being in the field of view another 64 registration experiments were conducted using an alternative 22 cm 3DRA reconstruction, which contained the facial structures. The average results of all phantom experiments and their standard variations are listed in Table III.

TABLE III
RESIDUAL ERROR AFTER REGISTRATION CHARACTERISTICS OF THE PHANTOM EXPERIMENTS. ROTATION IN DEGREES, TRANSLATION IN MILLIMETERS.
SUCCESSFUL EXPERIMENTS ARE THOSE THAT MEET THE CRITERIA (TRANSLATION ERROR < 1 MM, ROTATION ERROR < 3°)

|  | Results for all measurements | | | | Results for clinically relevant measurements | | | |
| --- | --- | --- | --- | --- | --- | --- | --- | --- |
|  | mean | stddev | min | max | mean | stddev | min | max |
| Rotation (all) | 1.330 | 2.646 | 0.010 | 42.601 | 0.952 | 1.852 | 0.010 | 22.284 |
| Rotation (successful) | 0.659 | 0.686 | 0.010 | 5.630 | 0.590 | 0.596 | 0.010 | 4.572 |
| Translation (all) | 1.519 | 3.204 | 0.005 | 33.068 | 0.657 | 2.238 | 0.005 | 33.068 |
| Translation (successful) | 0.283 | 0.211 | 0.005 | 1.410 | 0.276 | 0.218 | 0.005 | 1.410 |

*B. Clinical Data*

To test the registration on clinical data, we registered projection images of a rotational run with their volumetric reconstruction for ten patients. The data concerned intracranial 3DRA acquisitions during neurovascular procedures with image format diameters of 22 cm (6 patients) and 27 cm (4 patients). A translational and rotational offset was induced prior to the registration. To examine the influence of the vasculature being visible in the runs, we used five runs with iodine contrast medium injected in the vessels, while the other five runs were without. A random projection image was selected in the rotational run of 120 images. The translational component had a random length in the range of [0, 25] mm in a random direction. The distance of the translational offset was 12.54 mm on average (10.28 mm in the plane of the projection image and 6.09 mm perpendicular), with a standard deviation of 6.79 mm (6.07 mm within the projection plane and 4.91 mm perpendicular). The rotational offset amounted between 0 and 10 degrees around a random axis (mean 5.19°, stddev 2.79°). For each of the 10 patients 100 registrations were executed with each random image number, translational and rotational offset in the ranges described above.

The residual error after registration is presented in Fig. 7. 87.1% of the registration attempts passed the success criterion of a residual translational error of < 1 mm, and 83.9% passed the criterion of a rotational error of < 3°. The in-plane residual translational error amounted 0.89 mm on average (stddev 2.79 mm), and the mean rotational error was 2.63° (stddev 4.76°). When only the runs that pass the success-criteria are taken into account the in-plane translational errors had a mean of 0.24 mm (stddev 0.21 mm) and the average rotational error was 1.11° (stddev 0.68°).

The results can be examined separately for the contrast enhanced runs and contrast-less runs. 92.4% of the contrast enhanced runs passed the translational success criterion and 86.6% the rotational criterion. For the runs without contrast 81.8% passed the translational criterion and 81.2% the rotational criterion.

IV. DISCUSSION

The automatic image-based registration is of great clinical benefit during 3D roadmap navigation in minimal invasive treatment of neuro-vascular lesions. Patient motion is compensated in the 3D roadmap within moments after its occurrence. No manual interaction of the physician is required, which means that the physician can focus on the procedure rather on the interaction with the equipment. To indicate that automatic motion compensation has been applied, a message is shown on top of the image. When the 3DRA is registered to a multi-modality dataset (e.g., CT or MR), the automatic motion compensation can also be applied to an overlay of the multi-modal data [9,20,33].

The clinically relevant subset (see Table II) excludes the smallest image sizes of 15 and 19 centimeters, both for the volumetric reconstructions and the overlay images, since these are disabled in the commercially available software due to lacking usable landmark features. Furthermore, overlaying a small 3D reconstruction with a larger overlay image has also been excluded from the subset, since this leads to a limited 3D overlay field of view. Within the scope of this article, the results have been evaluated for the entire range and the clinically relevant subset. In the commercial product, though, the image-based registration is only enabled for the clinically relevant subset and uses the machine-based registration only outside this subset.

There was no significant difference between the residual offset of the phantom experiments that used fluoroscopy images and exposure images. Much more important is the presence of pronounced landmark features, like the facial structures and sinuses.

It should be noted that next to the described roadmapping application, there are a number of other application in which the described patient motion compensation can be (and is) applied; A related application is e.g. the navigation of a needle to a skull base tumor [34]. In applications that quantify the flow velocity and volume of contrast medium through the cerebral vasculature and aneurysms, a 3DRA dataset is registered to the 2D flow images to determine vessel foreshortening and vessel diameter [35,36]. Also, the 2D-3D registration can be used to back-project a tracked catheter [37] onto the 3D vessel tree. However, when applying that outside the cerebral domain, an elastic registration [38] may be required to account for non-rigid deformation.

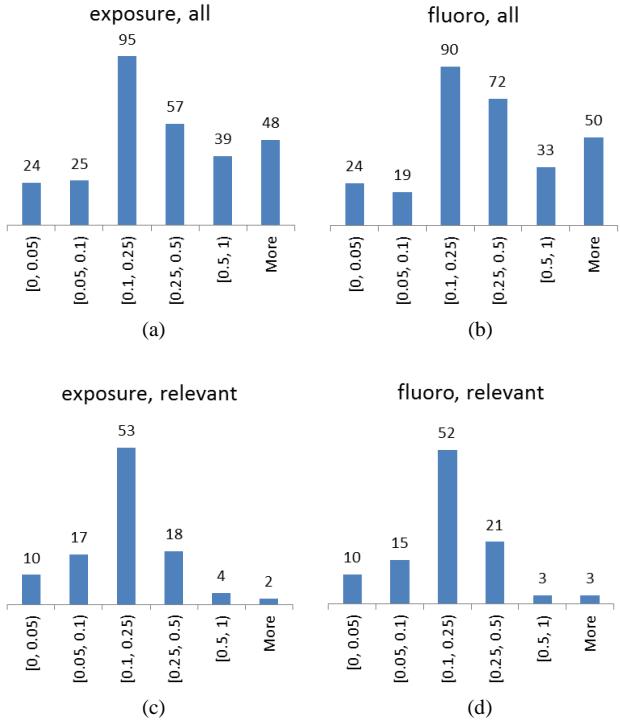

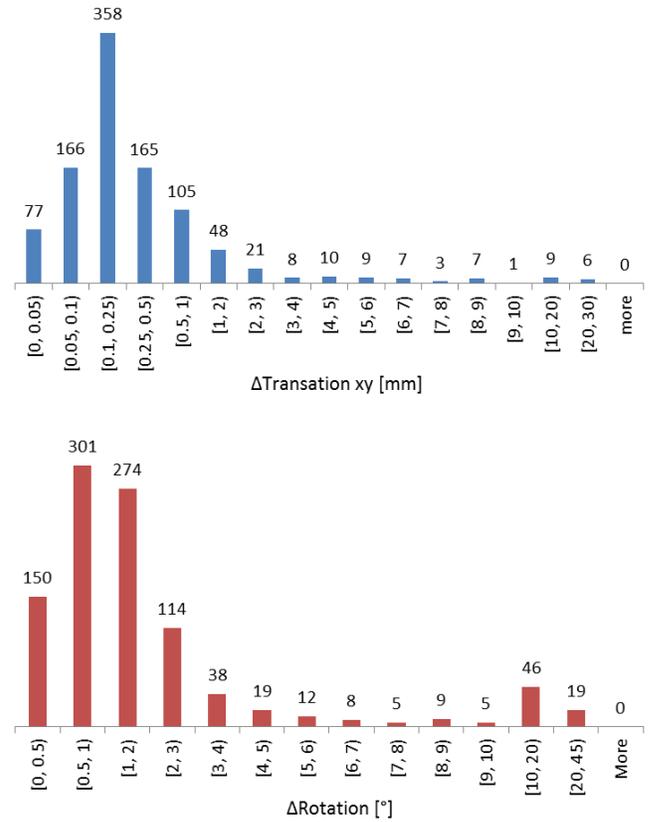

Fig. 6. Distribution of the translational residual error for the phantom experiments. The horizontal axis expresses the error ranges in mm and the bars indicate the number of results within a given range. The distribution over the successful range (< 1 mm) is shown and the amount of results outside the success criterion are specified in the "More" bin. Note that the intervals for the bins at the left are smaller. All experiments with exposure runs are reported in (a), all fluoroscopy experiments in (b), the clinically relevant subset for the exposure runs in (c), and the clinically relevant subset for the fluoroscopy runs in (d).

Fig. 7. Distribution of the translational and rotational residual error for the experiments using clinical data. The horizontal axis expresses the error ranges and the bars indicate the number of registration results within a given range. Note that the intervals for the bins at the left are smaller.

## V. CONCLUSIONS

The fusion of the live fluoroscopy image stream and a 3D rotational reconstruction of the vessel tree, known as "3D Roadmap", allows to reduce the usage of toxic iodine contrast medium, since contrast needs to be injected only during the rotational acquisition. The 3D roadmap requires the fluoroscopy images and the 3D reconstruction to be registered. Since this registration is being used during navigation of the endovascular devices, the possibilities for manual interaction are very limited. Typically, the registration is based on the sensor information of the X-ray C-arm regarding its geometry position. This kind of machine-based registration allows to change the C-arm viewing incidence angles, source-detector distance, table position, etc., while the registration is updated instantaneously. Since the fluoroscopy images and the 3D reconstruction are acquired with the same C-arm equipment, this is enough to provide a very accurate registration, provided that there was no patient motion. Any patient motion, however, will lead to a misregistration, which is particularly disturbing when working with very small devices.

This article has investigated the accuracy, robustness, and computation time aspects of an image-based registration algorithm that is applied after the machine-based registration, in order to correct for rigid patient motion in neuro-vascular minimally invasive treatment. The image-based registration is applied during the live guidance and does not require any manual interaction or initialization. The image-based registration has been implemented in a widely available commercial angiographic C-arm system and is used in clinical procedures on a daily basis.